%% file: acl_latex.tex
\documentclass[11pt]{article}

\usepackage[final]{acl}

\usepackage{times}
\usepackage{latexsym}

\usepackage[T1]{fontenc}

\usepackage[utf8]{inputenc}

\usepackage{microtype}

\usepackage{inconsolata}

\input{math_command}

\usepackage{graphicx}
\usepackage{microtype}
\usepackage{hyperref}
\usepackage{url}
\usepackage{booktabs}

\usepackage{algorithm}  
\usepackage{algorithmicx}  
\usepackage[noend]{algpseudocode}
\usepackage{amsmath}  
\usepackage{amsthm}
\usepackage{amsfonts}  
\usepackage{amssymb}  
\usepackage{graphicx}  
\usepackage{xcolor}
\usepackage{multirow}
\usepackage{xurl}
\usepackage{enumitem}
\usepackage{lineno}
\usepackage{makecell}
\usepackage{wrapfig}
\usepackage{subcaption}
\usepackage[utf8]{inputenc}

\newcommand{\eg}{\textit{e.g.}, }
\newcommand{\ie}{\textit{i.e.}, }

\newtheorem{theorem}{Theorem}[section]

\newtheorem{remark}[theorem]{Remark}
\newtheorem{proposition}[theorem]{Proposition}
\newtheorem{corollary}[theorem]{Corollary}
\title{Online Difficulty Filtering for Reasoning Oriented Reinforcement Learning}

\author{\textbf{Sanghwan Bae$^1$\thanks{Equal contribution.} \ \ Jiwoo Hong$^{2*}$\thanks{Work was done while the authors were at NAVER Cloud.} \ \ Min Young Lee$^1$ \ \ Hanbyul Kim$^1$ \ \ JeongYeon Nam$^{3}$\footnotemark[2]} \ \ \vspace{0.05in} \\  \textbf{Donghyun Kwak$^1$} \\
\\
NAVER Cloud$^1$ \ \ KAIST AI$^2$ \ \ TwelveLabs$^3$
\\
\footnotesize{\texttt{baaesh10@gmail.com}, \texttt{jiwoo\_hong@kaist.ac.kr}}
}

\begin{document}
\maketitle

\input{sections/0_abstract}
\input{sections/1_intro}
\input{sections/2_related_works}
\input{sections/3_prelim}
\input{sections/4_learnability}

\input{sections/5_method}

\input{sections/6_exps}
\input{sections/9_analysis}

\input{sections/8_conclusion}
\input{sections/limitations}

\bibliography{custom}

\input{sections/apdx}

\end{document}

%% file: math_command.tex
\usepackage{amsmath,amsfonts,bm}

\def\eqref#1{equation~\ref{#1}}

\def\1{\bm{1}}

\DeclareMathAlphabet{\mathsfit}{\encodingdefault}{\sfdefault}{m}{sl}
\SetMathAlphabet{\mathsfit}{bold}{\encodingdefault}{\sfdefault}{bx}{n}



%% file: sections/0_abstract.tex
\begin{abstract}

Recent advances in reinforcement learning with verifiable rewards (RLVR) show that large language models enhance their reasoning abilities when trained with verifiable signals. However, due to reward sparsity, effectiveness depends heavily on selecting samples of appropriate difficulty. In this work, we present a formal analysis of online difficulty-aware filtering and establish its theoretical foundations. We show that expected policy improvement is lower-bounded by the variance of task-level success probabilities, implying that selecting tasks of intermediate difficulty maximizes learning efficiency. Building on this, we demonstrate that balanced filtering maximizes this lower bound, leading to superior performance and sample efficiency. Evaluations across multiple math reasoning benchmarks validate that balanced filtering consistently enhances convergence speed and final performance, achieving up to +12\% gains in less than half the training steps of standard GRPO. By extending our analysis to various reward distributions, we provide a principled foundation for future RLVR curriculum strategies, confirmed through both theoretical analysis and extensive empirical results.
\end{abstract}

%% file: sections/1_intro.tex
\section{Introduction}

Reinforcement learning (RL) has become a key training paradigm for training large language models (LLMs) to further enhance their generational capabilities \citep{ouyang2022training,touvron2023llama2openfoundation,yang2024qwen2technicalreport,olmo20252olmo2furious}, often posed as \emph{post-training}. Specifically, reinforcement learning with verifiable rewards (RLVR) that have a discrete set of correct answers for domains like math reasoning is emerging as a new application of RL \citep{openai2024openaio1card,lambert2025tulu,guo2025deepseek}. Despite its effectiveness, the training efficiency is the main bottleneck in RL for LLMs, which recent works try to overcome either by hardware optimization \citep{mei2025real,noukhovitch2025faster} or algorithmic solutions \citep{lee2024towards,ahmadian2024back}.

Efficient learning, \ie achieving the optimal performance with less data, has long been studied in the education domain, where theories such as the Zone of Proximal Development \citep[ZPD]{vygotsky1978mind, tzannetos2023proximal} emphasize that learning is most efficient when tasks are \emph{neither too easy nor too hard}, but instead fall within a learner’s optimal challenge zone. This has motivated a variety of strategies in general language modeling \citep{platanios2019competence,maharana2022curriculum,xie2023data}. When the idea of filtering the data with intermediate difficulty is applied to RLVR, it can be used for progressively introducing harder problems \citep{team2025kimi} or filtering examples based on the pre-defined proxies \citep{muennighoff2025s1, ye2025limo,yang2025qwen3technicalreport}. Especially, by setting the proxy as the training policy's capability, we can harness the online learning in RLVR for difficulty-aware efficient learning \citep{cui2025process}. Despite applying difficulty-aware data curation can be found in recent works for RLVR, they often lack detailed analysis on \emph{how the theory of ZPD can be linked to online reinforcement learning algorithms}, \ie theoretical foundation of online difficulty-aware filtering in RLVR.
\begin{figure*}[t!]
    \centering
    \includegraphics[width=\textwidth]{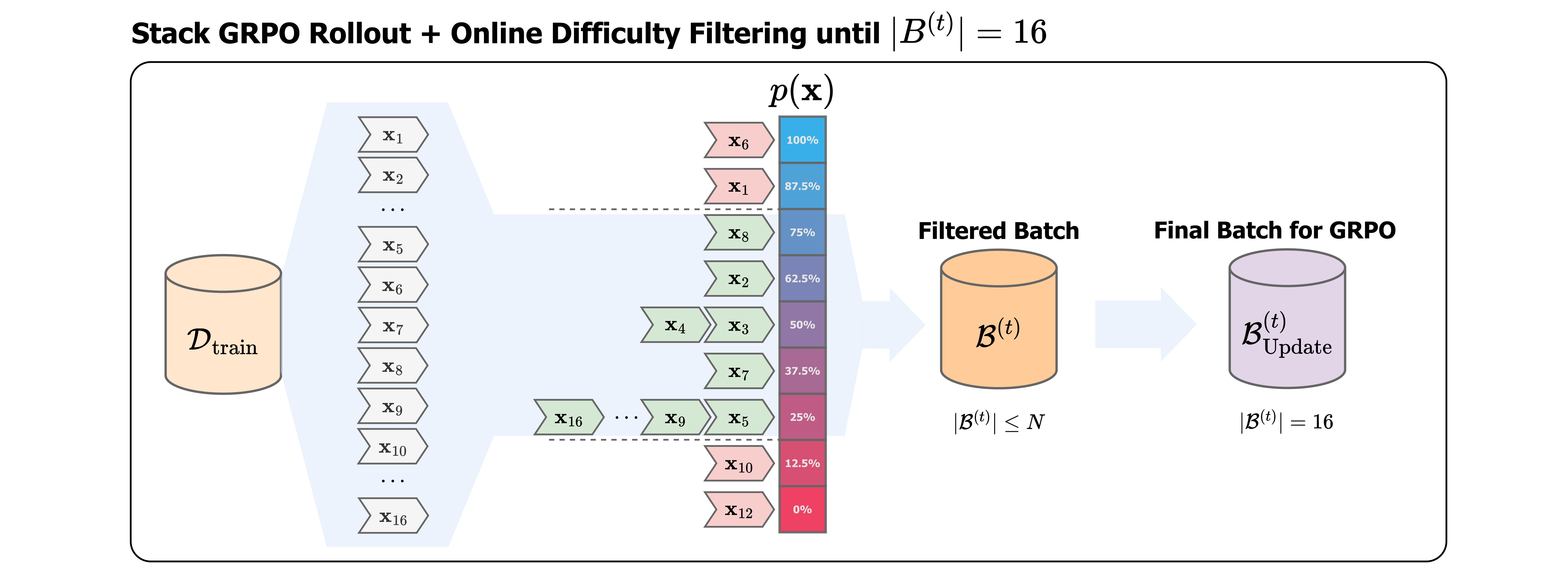}
    \caption{\textbf{Balanced online difficulty filtering} for maximizing the effectiveness of GRPO. With $G$ rollouts for each prompt $\mathbf{x}$, we measure the pass rate $p(\mathbf{x})$ as the average accuracy and filter them by predefined thresholds: \eg $0.25 \leq p(\mathbf{x}) \leq 0.75$ in this case. We recursively stack filtered prompts until the train batch size meets the fixed size $N$. We elaborate on the asynchronous implementation in Appendix \ref{apdx:implementation}.}
    \label{fig:main_figure}
\end{figure*}

In this work, we establish a theoretical foundation for online difficulty-aware filtering in RLVR through identifying the sample reward variance as the lower bound of the reversed KL divergence between the initial policy and the optimal policy, thereby being an effective proxy for filtration. With a novel asynchronous sampling strategy that replaces filtered-out items with parallel rollouts and retains the batch size, we build a theoretical and empirical background a reliable filtering strategy for RLVR. Our contributions are as follows:
\begin{enumerate}[left=1pt,parsep=2pt]
    \item \textbf{Theoretical assessment of prompt-level learnability} (\S\ref{sec:learnability}): We prove that the learning signal for the prompt, given the initial policy, can be approximated via sample reward variance (Proposition \ref{prop:variance_controlled}), implying why the prompts with extremely low or high pass rate should be filtered (Remarks \ref{remark:soft} and \ref{remark:hard}).
    \item \textbf{Empirical scalability and generalizability of balanced online difficulty filtering} (\S\ref{sec:result}): With 3B and 7B models, we empirically validate the effectiveness of the balanced online difficulty filtering across five math reasoning benchmarks with varying levels, \eg $+10\%$ on AIME and $+4.2\%$ in average for 3B and $+12\%$ on AMC and $+4.5\%$ in average for 7B.
    \item \textbf{Sample efficiency of balanced online difficulty filtering} (\S\ref{sec:analysis}): We show that the filtering in fact decreases the amount of training data and time in achieving the optimal performance, using less than half of gradient updates to outperform the plain GRPO with filtering.
    \item \textbf{Generalizability of sample reward variance as learnability proxy} (\S\ref{subsec:general_rewards}): We provide a general proof that wider range of reward distributions, \eg Gaussian (Corollary \ref{cor:gaussian_exact}) or Multinomial (Corollary \ref{cor:multinomial_lower}), can also enjoy the sample reward variance as the learnability proxy.
\end{enumerate}

%% file: sections/3_prelim.tex
\section{Preliminaries}


\paragraph{Reinforcement learning in language models.} Given the training policy $\pi_\theta$ initialized from the reference policy $\pi_\mathrm{init}$, reinforcement learning (RL) in language model environment optimizes $\pi_\theta$ to maximize the reward assessed by the reward function $r$ \citep{christiano2017deep,ziegler2020finetuning}:
\begin{equation}
    \max_\theta \mathbb{E}_{\mathbf{y}\sim \pi_\theta(\cdot | \mathbf{x})} \left[ r(\mathbf{x},\mathbf{y}) \right] - \beta  \mathbb{D}_\mathrm{KL}\left( \pi_\theta \| \pi_\mathrm{init}\right),\label{eq:obj}
\end{equation}
penalizing excessive divergence of $\pi_\theta$ with hyperparameter $\beta$ for the input and output token sequences $\mathbf{y} = \{ y_i \}_{i=1}^K$ and $\mathbf{x} = \{ x_i \}_{i=1}^M$. The policy gradient methods like REINFORCE \citep{williams1992simple} or PPO \citep{schulman2017proximal} are often applied, defining \emph{token-level} reward with the per-token divergence as a final reward \citep{ziegler2020finetuning, huang2024the}:
\begin{equation}
    r(\mathbf{x}, \mathbf{y}) - \beta \log \frac{\pi_\theta(\mathbf{y}|\mathbf{x})}{\pi_\mathrm{init}(\mathbf{y}|\mathbf{x})}.\label{eq:reward}
\end{equation} 
The corresponding optimal policy $\pi^*$ is well known to be defined with respect to $\pi_\mathrm{init}$ as \citep{korbak2022on, go2023aligning, rafailov2023direct}, 
\begin{equation}
    \pi^*(\mathbf{y}|\mathbf{x}) = Z(\mathbf{x})  \pi_{\mathrm{init}}(\mathbf{y}|\mathbf{x}) e^{\frac{1}{\beta}r(\mathbf{x}, \mathbf{y})},\label{eq:optim}
\end{equation}
where $Z(\mathbf{x})$ is the partition function that normalizes the action probability given $\mathbf{x}$.

\paragraph{Group relative policy optimization.} Unlike PPO, recent works exclude parameterized value models \citep{ahmadian2024back, kazemnejad2024vineppo, wu2024pairwise}, including group relative policy optimization \citep[GRPO]{shao2024deepseekmath}. GRPO leverages the PPO-style clipped surrogate objective but calculates the policy gradient by weighting the log-likelihood of each trajectory with its advantage, thus removing the need for a critic \citep{vojnovic2025alignmentobjectivegrpo, mroueh2025reinforcementlearningverifiablerewards}. For each prompt, \(G\) sampled responses and their reward $r_i$ is used to calculate the advantage $\hat{A}_i$:
\begin{equation}
    \hat{A}_i = \frac{r_i - \mathrm{mean}(r_1,\dots,r_G)}{\mathrm{std}(r_1,\dots,r_G)},\label{eq:grpo_adv}
\end{equation}
where $\mathrm{mean}(\cdot)$ and $\mathrm{std}(\cdot)$ are the average and standard deviation of the input values. The effectiveness of GRPO is especially highlighted in the tasks with verifiable reward stipulated through the binary reward functions \citep{lambert2024t, guo2025deepseek, wei2025swerladvancingllmreasoning}:
\begin{equation}
    r_\mathrm{acc}(\mathbf{x}, \mathbf{y}) = \begin{cases}
        ~~1 & \quad \text{if output is correct}\\ 
        ~~0 &\quad \text{otherwise.}
    \end{cases}\label{eq:acc_reward}
\end{equation}

%% file: sections/4_learnability.tex
\section{Learnability in GRPO and Online Difficulty Filtering}\label{sec:learnability}

We study the \emph{learnability} of prompts in reinforcement learning with language model environments under binary rewards. Our analysis shows that prompts that are either trivially easy or impossibly hard yield zero divergence and thus no learning signal, while intermediate prompts with higher reward variance maximize the effective gradient information. These results are formalized in Proposition~\ref{prop:variance_controlled}, which motivates a \textbf{balanced online difficulty filtering} strategy (\S\ref{subsec:method}-\S\ref{subsec:strategy}) for optimizing GRPO training.

\subsection{Background}\label{subsec:background}

The optimal value function and the partition function in the soft RL setting \citep{schulman2018equivalencepolicygradientssoft, richemond2024offlineregularisedreinforcementlearning} are defined as:
\begin{gather}
    V^*(\mathbf{x}) := \beta \log \mathbb{E}_{\mathbf{y}\sim \pi_\mathrm{init}(\cdot|\mathbf{x})}\left[ e^{\frac{1}{\beta} r\left(\mathbf{x}, \mathbf{y} \right)} \right]\\Z(\mathbf{x}) = \exp\left(\frac{1}{\beta}V^*(\mathbf{x})\right).
    \label{eq:value}
\end{gather}
Using \(V^*(\mathbf{x})\) in \eqref{eq:optim}, the log ratio between $\pi_\mathrm{init}$ and \(\pi^*\) can be expressed as:
\begin{equation}
    \log \frac{\pi^*(\mathbf{y}|\mathbf{x})}{\pi_{\mathrm{init}}(\mathbf{y}|\mathbf{x})} =  \frac{1}{\beta}\Big( r(\mathbf{x}, \mathbf{y}) - V^*(\mathbf{x}) \Big).
\end{equation}
Taking the expectation with respect to \(\pi_\mathrm{init}\) yields:
\begin{equation}
\begin{split}
    &\mathbb{E}_{\mathbf{y}\sim \pi_\mathrm{init}(\cdot|\mathbf{x})} \left[ \log \frac{\pi^*(\mathbf{y}|\mathbf{x})}{\pi_\mathrm{init}(\mathbf{y}|\mathbf{x})} \right]\\&= \frac{1}{\beta}\mathbb{E}_{\mathbf{y}\sim \pi_\mathrm{init}(\cdot|\mathbf{x})}\left[r(\mathbf{x}, \mathbf{y})\right] - \frac{1}{\beta}V^*(\mathbf{x}),
    \label{eq:logratio}
\end{split}
\end{equation}
where the right-hand side (RHS) is a soft-RL variant of the advantage function scaled by $\beta^{-1}$ \citep{haarnoja2017reinforcement, schulman2018equivalencepolicygradientssoft}, as $\mathbb{E}_{\pi_\mathrm{init}}\left[ r(\mathbf{x}, \mathbf{y}) \right]$ can be interpreted as a Q-function. And the left-hand side (LHS) corresponds to the negative reverse KL divergence between \(\pi_\mathrm{init}\) and \(\pi^*\) \citep{rafailov2024from}:
\begin{equation}
\begin{split}
    &\mathbb{D}_\mathrm{KL}\left( \pi_\mathrm{init}(\mathbf{y}|\mathbf{x})|\pi^*(\mathbf{y} \,\|\, \mathbf{x}) \right)\\&~~= -\mathbb{E}_{\mathbf{y}\sim \pi_\mathrm{init}(\cdot|\mathbf{x})} \left[ \log \frac{\pi^*(\mathbf{y}|\mathbf{x})}{\pi_\mathrm{init}(\mathbf{y}|\mathbf{x})} \right].
\end{split}
\end{equation}

\paragraph{Learnability in binary reward case.} For the binary reward \(r_\mathrm{acc}\), the reward distribution is Bernoulli with parameter $p(\mathbf{x})$ for prompt $\mathbf{x}$, policy $\pi$, and $\mathbf{y} \sim \pi(\cdot | \mathbf{x})$, which we refer as ``\emph{pass rate}'':
\begin{equation}
    p(\mathbf{x}) = \mathbb{E}_{\pi_\mathrm{init}}\left[r_\mathrm{acc}(\mathbf{x}, \mathbf{y})\right],
    \label{eq:success}
\end{equation}
and variance \(p(\mathbf{x})(1-p(\mathbf{x}))\). Here, we categorize the prompts into five categories: 
\vspace{-0.01in}
\begin{enumerate}[left=4pt,itemsep=0.05mm]
    \item \textbf{Absolute-hard} ($\mathbf{x}_\mathrm{Hard},~p(\mathbf{x}_\mathrm{Hard}) = 0$)
    \item \textbf{Soft-hard} ($\mathbf{x}_\mathrm{hard},p(\mathbf{x}_\mathrm{hard}) = \epsilon$)
    \item \textbf{Intermediate} ($\mathbf{x}_\mathrm{inter},~\epsilon \leq p(\mathbf{x}_\mathrm{inter}) \leq 1 -  \epsilon$)
    \item \textbf{Soft-easy} ($\mathbf{x}_\mathrm{easy},~p(\mathbf{x}_\mathrm{easy}) = 1 - \epsilon$)
    \item \textbf{Absolute-easy} ($\mathbf{x}_\mathrm{Easy},~p(\mathbf{x}_\mathrm{Easy}) = 1$)
\end{enumerate}
\vspace{-0.01in}
where $\epsilon$ is a small positive constant satisfying $0 \ll \epsilon < 0.5$. The variance is zero if and only if \(p(\mathbf{x})=0\) or \(p(\mathbf{x})=1\), corresponding to \emph{absolute hard} and \emph{absolute easy} prompts, respectively.

\subsection{Prompt-level learnability: theoretical analysis}
\label{subsec:learnability}

We analyze the learnability of prompts under the soft reinforcement learning formulation defined in \S\ref{subsec:background}. Let the binary verifiable reward $r_{\mathrm{acc}}(x,y)\in\{0,1\}$ follow a Bernoulli distribution,
\begin{equation}
p(x) := \mathbb{E}_{y\sim\pi_{\mathrm{init}}(\cdot|x)}[r_{\mathrm{acc}}(x,y)],
\end{equation}
which we refer to as the \emph{pass rate} of prompt $x$. Given the definitions of the value function and log-ratio, the expected log ratio between $\pi_{\mathrm{init}}$ and the soft-optimal policy $\pi^*$ can be written as
\begin{equation}
\begin{split}
    &\mathbb{E}_{y\sim\pi_{\mathrm{init}}(\cdot|x)}
    ~\left[\log\frac{\pi^*(y|x)}{\pi_{\mathrm{init}}(y|x)}\right]
    \\&= \frac{p(x)}{\beta}
    - \log~\Big((1-p(x))+p(x)e^{1/\beta}\Big),
    \label{eq:bernoulli_logratio}
\end{split}
\end{equation}
and its negative corresponds to the reverse KL divergence
$\mathbb{D}_\mathrm{KL}\left(\pi_{\mathrm{init}}(\cdot|x)\,\|\,\pi^*(\cdot|x)\right)$
defined in \eqref{eq:logratio}.

\begin{proposition}[Variance-controlled separation and degeneracy]
\label{prop:variance_controlled}
For any prompt $x$ and temperature $\beta>0$, the reverse KL divergence between the initial and optimal policies, $\pi_{\mathrm{init}}$ and $\pi^*$, satisfies
\begin{equation}
    \mathbb{D}_\mathrm{KL}~\big(\pi_{\mathrm{init}}(\cdot|x)\|\pi^*(\cdot|x)\big)
    \;\ge\;
    \frac{p(x)\big(1-p(x)\big)}{2\beta^2},
    \label{eq:var_lower_bound}
\end{equation}
with the bound maximized at $p(x)=\tfrac12$. 
In particular, for absolute-hard or absolute-easy prompts where $p(x)\in\{0,1\}$, the divergence vanishes and $\pi_{\mathrm{init}}$ is already optimal:
\begin{equation}
    \mathbb{D}_\mathrm{KL}~\big(\pi_{\mathrm{init}}(\cdot|x)\,\|\,\pi^*(\cdot|x)\big) = 0.
\end{equation}
\end{proposition}
\noindent \textit{Proof sketch.} Starting from \eqref{eq:bernoulli_logratio} and expanding 
$\log~\big((1-p)+pe^{1/\beta}\big)$ 
with $\log(1+\epsilon)\ge\epsilon-\tfrac{\epsilon^2}{2}$ for $\epsilon=p(x)(1/\beta+1/(2\beta^2))$, we obtain
\begin{equation}
\mathbb{E}_{y\sim\pi_{\mathrm{init}}}\left[\log\frac{\pi^*(y|x)}{\pi_{\mathrm{init}}(y|x)}\right]
\le -\frac{p(x)\big(1-p(x)\big)}{2\beta^2}.
\end{equation}
Since the reverse KL divergence is the negative of this expectation \eqref{eq:logratio}, 
the inequality in \eqref{eq:var_lower_bound} follows. 
When $p(x)\in\{0,1\}$, both the expected reward and the soft value function coincide ($V^*(x)=r_{\mathrm{acc}}(x,y)$), yielding zero divergence. The complete proof is provided in Appendix~\ref{apdx:learnability}.
\hfill$\square$

\begin{remark}[Learnability in absolute prompts: No learning signal at extremes]
\label{remark:hard}
For absolute-hard or absolute-easy prompts ($p(x)\in\{0,1\}$), the advantage term in GRPO \eqref{eq:grpo_adv} becomes zero for all rollouts, 
indicating that such prompts provide no gradient signal during policy optimization.
\end{remark}

\begin{remark}[Learnability in soft prompts: Maximal learnability band]
\label{remark:soft}
The lower bound in \eqref{eq:var_lower_bound} is proportional to the Bernoulli variance $p(x)(1-p(x))$, which attains its maximum at $p(x)=0.5$. 
Hence, prompts whose pass rates lie in the intermediate region $\epsilon \le p(x) \le 1-\epsilon$ provide the strongest learning signal, 
while soft-hard ($p(x)\approx\epsilon$) and soft-easy ($p(x)\approx1-\epsilon$) prompts contribute only marginally.
\end{remark}
Proposition~\ref{prop:variance_controlled} unifies both extreme and intermediate cases of prompt learnability. 
When $p(x)$ approaches the extremes, the reverse KL divergence $\mathbb{D}_\mathrm{KL}~\big(\pi_{\mathrm{init}}\|\pi^*\big)$ tends to zero, implying no separation between policies and thus no effective update. Conversely, when $p(x)\approx0.5$, the reward variance, \ie learnability, is maximized. These insights directly motivate the filtering criterion in our online curriculum, which selectively retains prompts within the intermediate difficulty band to ensure the highest learning efficiency.

%% file: sections/5_method.tex
\begin{algorithm*}[t!] 
\caption{Iterative GRPO with Online Difficulty Filtering}
\label{alg:grpo_async}
\small
\begin{algorithmic}[1]

\Require 
  Initial policy model $\pi_{\mathrm{init}}$; 
  Reward $r$; 
  Prompts queue $\mathcal{Q}$; 
  Pass rate thresholds $T_{\text{Low}}, T_{\text{High}}$; 
  Batch size $N$; 
  Group size $G$;
  $r_{\text{acc}}$ \eqref{eq:acc_reward};
  Visit count $\mathrm{vc}(\mathbf{x})$.
  
\State $\mathcal{P}_{\text{active}}$: The set of examples currently undergoing asynchronous rollout.
\State $C_{\text{max}}$: The maximum number of examples that can be processed concurrently.

\Function{$f_{async}$}{$\mathbf{x}$}
  \State $\,\{\mathbf{y}_{i}\}_{i=1}^{G} \sim \pi_{\theta}(\cdot \mid \mathbf{x})$
  \If{$\,T_{\text{Low}} \,\le\, \frac{1}{G} \sum_{i=1}^{G} r_{acc}(\mathbf{x},\mathbf{y}_i) \,\le\, T_{\text{High}}$}
    \State $\mathcal{B}^{(t)} \gets \mathcal{B}^{(t)} \cup \left\{(\mathbf{x}, \{\mathbf{y}_i\}_{i=1}^{G}, \{r(\mathbf{x}, \mathbf{y}_i)\}_{i=1}^{G})\right\}$
  \EndIf
  \State $\mathrm{vc}(\mathbf{x}) \gets \mathrm{vc}(\mathbf{x}) + 1$
\EndFunction

\State Initialize policy model $\pi_{\theta} \gets \pi_{\mathrm{init}}$
\State Initialize visit count $\mathrm{vc}(\mathbf{x}) \gets 0$ for all $\mathbf{x} \in \mathcal{D}$
\For{$\text{iteration} = 1, \ldots, I$}
  \State Initialize reference model $\pi_{\text{ref}} \gets \pi_{\theta}$
  \For{$\text{step} = 1, \ldots, M$}
    \State Initialize $\mathcal{B}^{(t)} \gets \varnothing$, $\mathcal{P}_{\text{active}} \gets \varnothing$
    \State Sort examples by visit count $\mathcal{Q} \gets \mathrm{sort}_{\mathrm{vc}}(\mathcal{D})$
    \While{$|\mathcal{B}^{(t)}| < N$}
      \If{$|\mathcal{P}_{\text{active}}| < C_{\text{max}}$}
        \State $\mathbf{x} \gets \text{nextExample}(\mathcal{Q})$
        \State $\mathcal{P}_{\text{active}} \gets \mathcal{P}_{\text{active}} \cup f_{async}(\mathbf{x})$ 
      \EndIf
    \EndWhile
    \State Cancel $\mathcal{P}_{\text{active}}$
    \State Compute $\hat{A}_{i}$ for $\mathbf{y}_i$ in $\mathcal{B}^{(t)}$ through group relative advantage estimation \eqref{eq:grpo_adv}.
    \State Update the policy model $\pi_\theta$ by maximizing the GRPO objective.
  \EndFor
\EndFor

\State \textbf{Output} $\pi_{\theta}$

\end{algorithmic}
\end{algorithm*}



\subsection{Method: online difficulty filtering with fixed batch size}\label{subsec:method}

From this vein, it is reasonable to comprise the input prompt set with \emph{intermediate} difficulty. Furthermore, balanced difficulty in the prompt set encourages balanced model updates for penalizing bad trajectories and reinforcing good trajectories in GRPO \citep{mroueh2025reinforcementlearningverifiablerewards}.

We analyze an online difficulty filtering approach that ensures a fixed batch size throughout training for a reasoning-oriented agent. Unlike static curricula with predefined difficulty orderings in problems \citep{yang2024qwen25mathtechnicalreportmathematical, team2025kimi, li2025limr}, our approach dynamically assesses difficulty \emph{on the fly} in each training step and applies difficulty filtering logic following the theoretical insights studied in \S\ref{sec:learnability}. 
We describe the detailed process in Algorithm \ref{alg:grpo_async} and the high-level illustration of the algorithm in Figure \ref{fig:high_level} in Appendix \ref{apdx:implementation}.

\paragraph{Online difficulty filtering with sample success rate for learnability.} First, we fill the batch $\mathcal{B}^{(t)}$ of the training step $t$ with filtered examples by measuring the success rate $p(\mathbf{x})$, \eqref{eq:success}, of each prompt $\mathbf{x}$ using sampled rollouts with size of $G$ as in \eqref{eq:batch}. With the predefined difficulty threshold $T_{\text{Low}}$ and $T_{\text{High}}$, we asynchronously filter and fill the batch to meet the fixed batch size.



\paragraph{Ensuring fixed batch size with asynchronous sampling and efficient batching.} While we showed that online difficulty filtering could maximize learnability in GRPO, naive filtering could result in inconsistent training batch size, leading to training instability and degraded performance \citep{li2022the}. For this reason, we ensure the fixed batch size to $|\mathcal{B}| = N$ for the batch $\mathcal{B}^{(t)}$ at the training step $t$ as in \eqref{eq:batch}. We extend technical details in Appendix \ref{apdx:implementation}.


\subsection{Difficulty filtering strategies}\label{subsec:strategy}

Based on the literature study in Appendix \ref{apdx:related}, we experiment two different difficulty filtering strategies, \textbf{balanced} and \textbf{skewed} difficulty filtering:
\begin{enumerate}[left=1pt]
    \item \textbf{Balanced difficulty filtering}: We set the thresholds to be symmetric to the success rate of $0.5$: \eg $T_\mathrm{High} = 0.8$ and $T_\mathrm{Low} = 0.2$.
    \item \textbf{Skewed difficulty filtering}: We set asymmetric thresholds, only filtering either easy or hard prompts: \eg $T_\mathrm{High} = 0.6$ and $T_\mathrm{Low} = 0$.
\end{enumerate}
We test if incorporating either side of extreme pass rate cases can boost the performance of online difficulty filtering, even though the learnability for either side has the same lower bound as in \S\ref{subsec:learnability}.

%% file: sections/6_exps.tex
\section{Experiments}\label{sec:exp}

\subsection{Experimental Setup}

\paragraph{Supervised fine-tuning.}Before reinforcement learning with verifiable rewards (RLVR) experiments, we fine-tune Qwen2.5-3B base \citep{yang2024qwen2} as a cold start, following \citet{guo2025deepseek}. Specifically, we curate 1.1K verified problem-solution pairs, with math problems sampled from NuminaMath \citep{li2024numinamath} and solutions distilled from DeepSeek-R1 \citep{guo2025deepseek}. 

\paragraph{Reinforcement learning with verifiable rewards.}For RLVR, we employ GRPO on top of the SFT checkpoint. In each training step, the model generates 16 rollouts for 16 prompts drawn from NuminaMath problems and receives a reward based on their correctness. We leave out 1,024 problems as a validation set. We also add a format reward and a language reward as in \citet{guo2025deepseek}. Additional training details are reported in the Appendix \ref{apdx:config}.

\input{table/merged}
\subsection{Experimental design}

\paragraph{Different strategies in online difficulty filtering.}Along with the plain GRPO without any prompt filtering, we test the online difficulty filtering with two different strategies introduced in \S\ref{subsec:strategy}: \ie balanced and skewed filtering. For the balanced setting, we test $(T_\mathrm{Low}, T_\mathrm{High}) \in \left\{ (0, 1), (0.1, 0.9), (0.2, 0.8), (0.3, 0.7), (0.4, 0.6) \right\}$. For a skewed setting, we sweep $T_\mathrm{Low}$ in $\left\{ 0, 0.2, 0.4\right\}$ when $T_\mathrm{High} = 1$ and $T_\mathrm{High}$ in $\left\{ 0.6, 0.8, 1\right\}$ when $T_\mathrm{Low} = 0$.
\paragraph{Comparison against existing offline filtering methods.} We mainly compare two offline difficulty filtering methods with our approach: offline data curation \citep{yang2024qwen25mathtechnicalreportmathematical,cui2025process,muennighoff2025s1,ye2025limo} and offline scheduling \citep{team2025kimi,li2025limr}. Offline data curation refers to the strategy that filters the problems by their difficulty before training, and offline scheduling additionally orders the training batches accordingly. For both offline strategies, we used Qwen2.5-7B-Instruct \citep{yang2024qwen2} or our SFT model as the difficulty proxies.

\paragraph{Evaluation Benchmarks.} We evaluate pass@1 across math reasoning benchmarks of varying difficulty levels: MATH500 \citep{hendrycks2measuring}, AIME \citep{li2024numinamath}, AMC \citep{li2024numinamath}, MinervaMath \citep{lewkowycz2022solving}, and OlympiadBench \citep{he2024olympiadbench} (See Appendix \ref{apdx:eval_benchmarks}).


\section{Results}\label{sec:result}

We first compare different online filtering strategies in \S\ref{subsec:online_exp} and expand to existing offline difficulty filtering methods in \S\ref{subsec:offline_exp}.

\subsection{Balanced and skewed filtering}\label{subsec:online_exp}

\paragraph{Balanced online difficulty filtering consistently outperforms plain GRPO.}
In Table \ref{tab:main_result}, balanced filtering (``Balanced'') outperforms the plain GRPO (``Plain'') on the average score of five challenging math reasoning benchmarks in all five threshold choices.
While fine-tuning the SFT checkpoint with plain GRPO without filtering reaches an average score of 26.3\%, balanced filtering achieves over 30\%, with overall improvements across the benchmarks.
For instance, balanced filtering achieved up to 10\% point improvement in AIME, which is the most difficult benchmark as shown through the accuracy in Table \ref{tab:main_result}. This supports our theoretical analysis in \S\ref{sec:learnability}, as online difficulty filtering enhances the effectiveness of GRPO training compared to the plain version without any filtering.

\input{table/scalability}
\paragraph{Progressively stricter threshold in balanced filtering incrementally improves performance.} 
By tightening the pass rate threshold $(T_\mathrm{Low}, T_\mathrm{High})$ for balanced filtering in Table \ref{tab:main_result}, the average score of five benchmarks starts from 27.3\% in $(0, 1)$, gradually increasing until over 30\% in $(0.3, 0.7)$.
Furthermore, simply removing examples in $(0, 1)$ that do not contribute to learning in GRPO results in a slight improvement over the baseline, aligning with Remark \ref{remark:hard}, \ie $\hat{A}$ is zero for $(0, 1)$. This result suggests that excluding ineffective examples improves both performance and training efficiency by focusing updates on meaningful data.
These observations are further supported by the difficulty-level analysis using MATH500 that provides five different levels in Appendix~\ref{apdx:difficulty}, which shows consistent gains across different levels.

\paragraph{Skewed online difficulty filtering is less effective than plain GRPO.}
While skewed filtering (``Skewed'') in Table \ref{tab:main_result} improves average performance up to 5.7\% over the SFT checkpoint, plain GRPO with 26.3\% outperforms skewed filtering consistently in every threshold choice, which achieves around 24.9\% to 25.9\%. Overall, maximizing the expected \emph{learnability} in GRPO enhances learning in complex reasoning tasks. As discussed in \S\ref{subsec:method}, balanced filtering emerges as the best choice since it balances between penalizing and reinforcing diverse explorations.

\subsection{Offline and online filtering}\label{subsec:offline_exp}

We apply the offline difficulty filtering with implementations from previous works \citep{yang2024qwen2}, with balanced threshold $(T_\mathrm{Low}, T_\mathrm{High}) = (0.2, 0.8)$ following the results in \S\ref{subsec:online_exp}.
While both offline curation (``Curation'') and offline scheduling (``Schedule'') in Table \ref{tab:main_result} show marginal improvements over plain GRPO with a maximum $2.1\%$ improvement, balanced online difficulty filtering consistently outperforms offline methods. Within offline methods, using an external difficulty assessment proxy (``External model'') exceeded the case using the SFT checkpoint (``initial model'') on average, but with varying results by benchmark.

\subsection{Scalability of balanced online filtering}

We adopt 7B scale model within the same Qwen2.5 family to confirm the scalability of the proposed method. In Table \ref{tab:scalability_result}, stricter filtering thresholds ($0.3 < p(\mathbf{x}) < 0.7$) yield the strongest performance with 3\% and 5\% increase in AIME and AMC, respectively. Overall, the ascending trend in Table \ref{tab:scalability_result} aligns with the 3B cases, demonstrating the scalability of online difficulty filtering.

%% file: table/merged.tex
\begin{table*}[t!]
    \centering
    \caption{Five math reasoning benchmark evaluation results with Qwen2.5-3B. ``Minerva.'' and ``Olympiad.'' refer to MinervaMath and OlympiadBench. ``External'' and ``Initial'' in offline filtering indicate using Qwen2.5-7B-Instruct and our SFT model as a difficulty proxy for filtering. $p(\mathbf{x})$ \eqref{eq:success} is the pass rate, the average correctness of rollouts. The highest and the second highest scores in each benchmark are highlighted with \textbf{bold} and \underline{underline}.}
    \resizebox{0.9\textwidth}{!}{
    \begin{tabular}{clcccccc}
        \toprule
         \textbf{Method} & \textbf{Difficulty Filter} & \textbf{MATH500} & \textbf{AIME} & \textbf{AMC} & \textbf{Minerva.} & \textbf{Olympiad.} & \textbf{Avg.} \\
        \midrule
        \makecell{\textbf{SFT}} & - & 49.8 & 0.0 & 20.5 & 13.2 & 17.3 & 20.2 \vspace{0.02in}\\
        \midrule
        \multirow{6}{*}{\makecell{\textbf{GRPO}\\\textbf{w/ Offline}\\\textbf{Filtering}}} 
         & \multicolumn{7}{l}{\textbf{Curation}}  \vspace{0.025in}\\
         & External model & 59.6 & 6.6  & 27.7 & 24.3 & 23.9 & 28.4 \\
         & Initial model  & 55.6 & \underline{10.0} & 28.9 & 18.8 & 18.2 & 26.3  \vspace{0.05in}\\
         & \multicolumn{7}{l}{\textbf{Schedule}}  \vspace{0.025in}\\
         & External model & 57.8 & \underline{10.0} & 28.9 & 20.6 & 21.5 & 27.8 \\
         & Initial model  & 57.0 & 3.3  & 28.9 & 19.1 & 24.9 & 26.7 \\
        \midrule
        \multirow{14}{*}{\makecell{\textbf{GRPO}\\\textbf{w/ Online}\\\textbf{Filtering}\\\textbf{(\textit{Ours})}}} 
         & \multicolumn{7}{l}{\textbf{Plain}}  \vspace{0.025in}\\
         & $0 \leq p(\mathbf{x}) \leq 1$ & 57.2 & 3.3 & 30.1 & 18.7 & 22.2 & 26.3 \vspace{0.05in}\\
         & \multicolumn{7}{l}{\textbf{Skewed}}  \vspace{0.025in}\\
         & $0 < p(\mathbf{x}) \leq 1$    & 57.0 & 0.0  & 26.5 & 19.8 & 21.4 & 24.9 \\
         & $0.2 < p(\mathbf{x}) \leq 1$   & 60.4 & 0.0  & 27.7 & 17.2 & 24.5 & 25.9 \\
         & $0.4 < p(\mathbf{x}) \leq 1$   & 55.8 & 0.0  & 21.7 & 19.9 & 21.6 & 23.8 \\
         & $0 \leq p(\mathbf{x}) < 1.0$   & 55.4 & 3.3  & 22.8 & 19.8 & 19.8 & 24.2 \\
         & $0 \leq p(\mathbf{x}) < 0.8$   & 56.2 & 0.0  & 28.9 & 17.2 & 21.7 & 24.8 \\
         & $0 \leq p(\mathbf{x}) < 0.6$   & 56.2 & 3.3  & 26.5 & 21.3 & 21.6 & 25.8 \vspace{0.05in}\\
         & \multicolumn{7}{l}{\textbf{Balanced}}  \vspace{0.025in}\\
         & $0 < p(\mathbf{x}) < 1$   & 60.8 & 3.3  & \underline{31.3} & 18.0 & \textbf{27.3} & 27.3 \\
         & $0.1 < p(\mathbf{x}) < 0.9$ & 58.8 & \textbf{13.3} & 25.3 & 22.4 & 22.2 & 28.4 \\
         & $0.2 < p(\mathbf{x}) < 0.8$ & \underline{62.2} & \underline{10.0} & 30.1 & 20.5 & \underline{26.3} & 29.8 \\
         & $0.3 < p(\mathbf{x}) < 0.7$ & \textbf{64.6} & 6.6  & 28.9 & \textbf{25.4} & 24.7 & \textbf{30.1} \\
         & $0.4 < p(\mathbf{x}) < 0.6$ & 60.2 & 6.6  & \textbf{32.8} & \underline{25.0} & 24.9 & \underline{29.9}\\
        \bottomrule
    \end{tabular}
    }
    \label{tab:main_result}
\end{table*}

%% file: table/scalability.tex

\begin{table*}[t!]
    \centering
    \small
    \caption{Five math reasoning benchmark evaluations with Qwen2.5-7B. The notations follow that of Table \ref{tab:main_result}.}
    \resizebox{0.8\textwidth}{!}{
    \begin{tabular}{clcccccc}
        \toprule
        \textbf{Method} & \textbf{Difficulty Filter} & \textbf{MATH500} & \textbf{AIME} & \textbf{AMC} & \textbf{Minerva} & \textbf{Olympiad} & \textbf{Avg.} \\
        \midrule
        \textbf{SFT} & – & 72.6 & 12.1 & 34.9 & 32.0 & 35.1 & 37.3 \\
        \midrule
        \multirow{5}{*}{\makecell{\textbf{GRPO}\\\textbf{w/ Online}\\\textbf{Filtering}\\\textbf{(Ours)}}}
         & \multicolumn{7}{l}{\textbf{Plain}}  \vspace{0.025in}\\
         & $0 \leq p(\mathbf{x}) \leq 1$       & \underline{75.0} & 12.3 & \underline{42.2} & 32.7 & 35.7 & 39.6 \vspace{0.05in}\\
         & \multicolumn{7}{l}{\textbf{Balanced}}  \vspace{0.025in}\\
         & $0 < p(\mathbf{x}) < 1$             & \underline{75.0} & \underline{13.1} & 41.0 & \underline{33.5} & \underline{36.9} & \underline{39.9} \\
         & $0.3 < p(\mathbf{x}) < 0.7$         & \textbf{75.8}    & \textbf{15.0}    & \textbf{47.0}    & \textbf{33.8}    & \textbf{37.6}    & \textbf{41.8} \\
        \bottomrule
    \end{tabular}
    }
    \label{tab:scalability_result}
\end{table*}

%% file: sections/9_analysis.tex
\section{Analysis}\label{sec:analysis}

\subsection{Learning dynamics analysis}

\begin{figure}[t!]
    \centering
    \includegraphics[width=\linewidth]{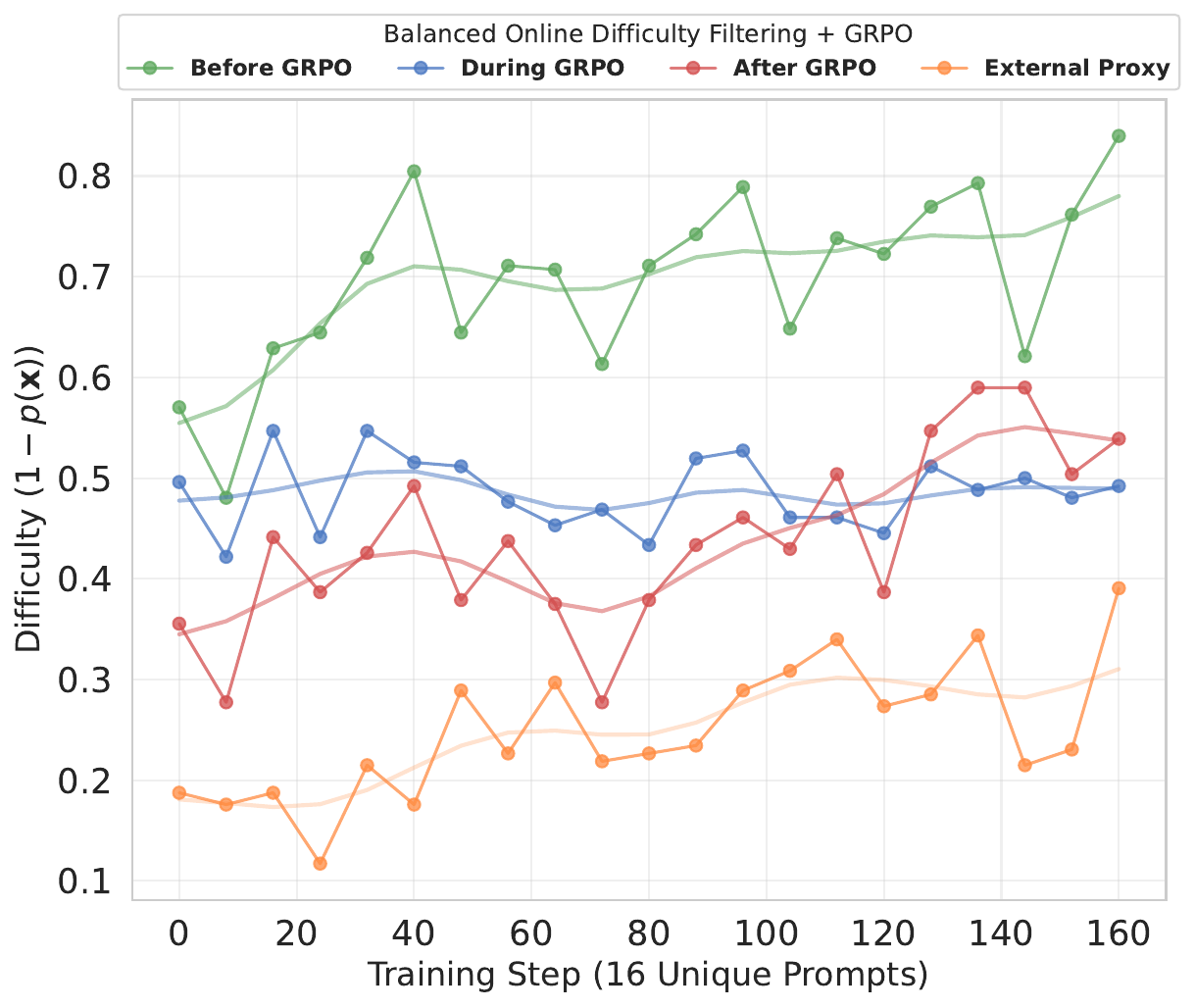}
    \caption{Perceived difficulty per batch curated through balanced online filtering. Defining ``difficulty'' as $1 - p(\mathbf{x})$, a greater difficulty implies lower sample accuracy.}
    \label{fig:difficulty}
\end{figure}

\begin{figure*}[t!]
    \centering
    \begin{subfigure}{0.48\linewidth}
        \includegraphics[width=\textwidth]{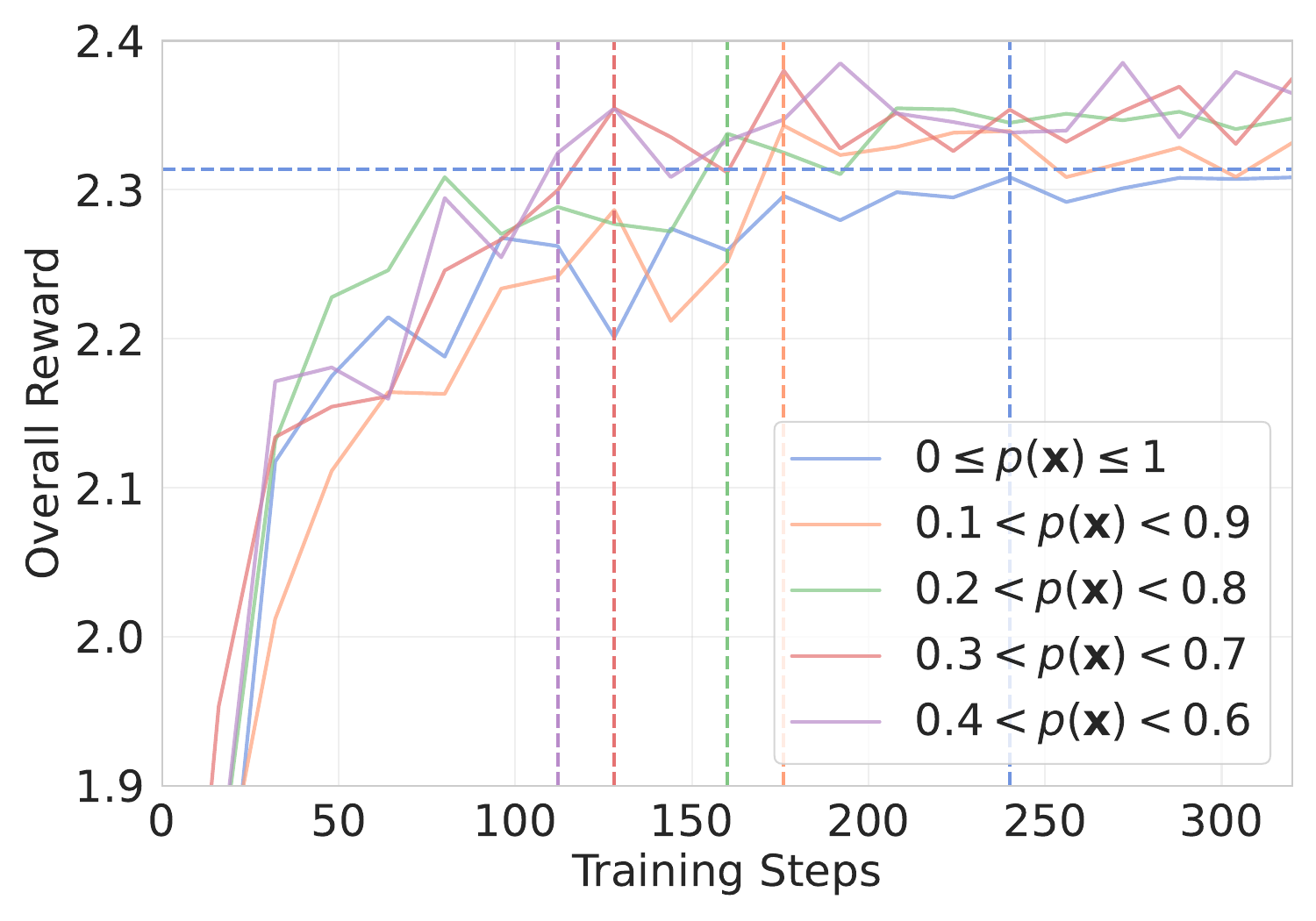}
        \caption{Validation Accuracy over Training Steps}
        \label{fig:step_}
    \end{subfigure}
    \begin{subfigure}{0.48\linewidth}
        \includegraphics[width=\textwidth]{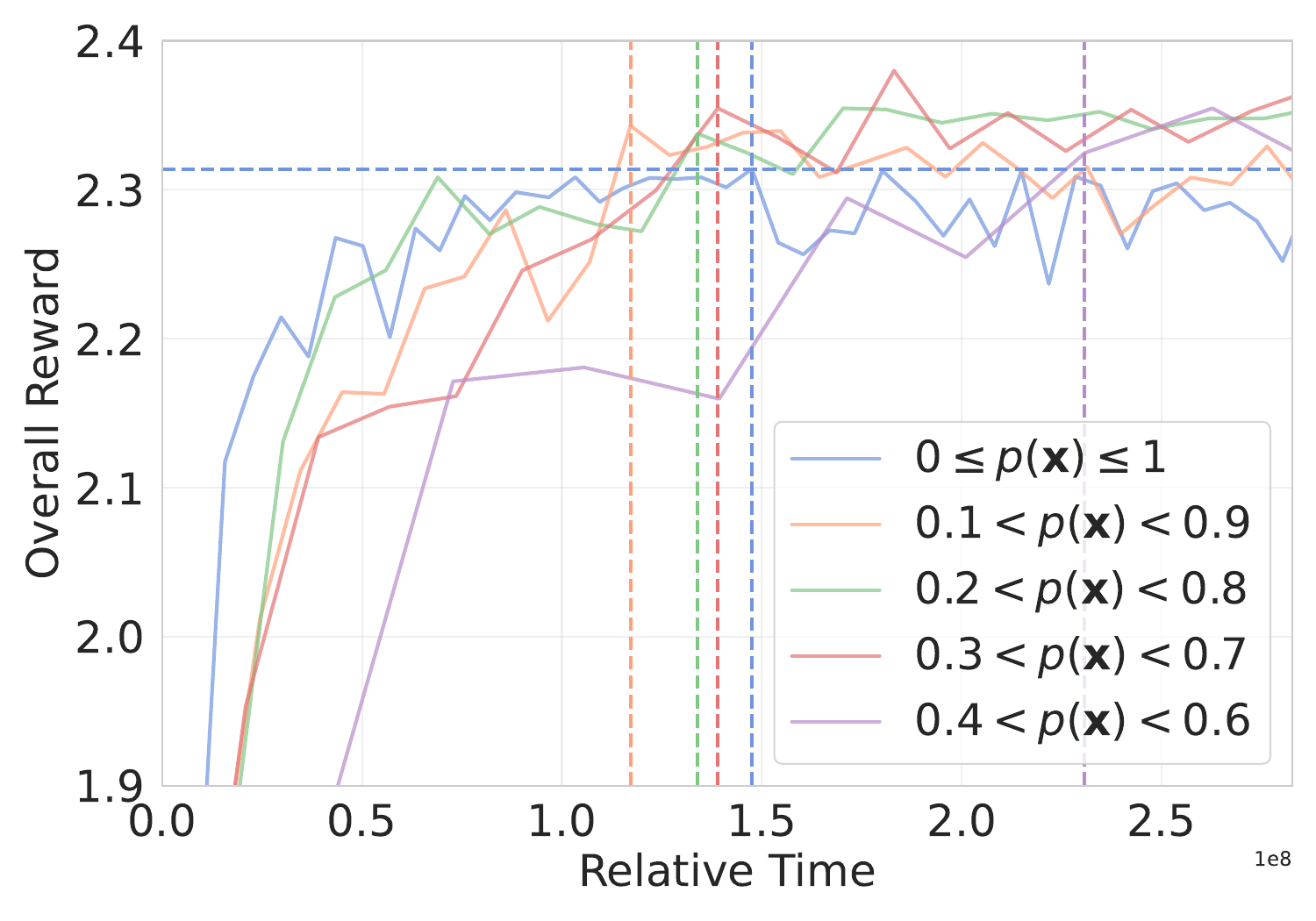}
        \caption{Validation Accuracy over Training Time}
        \label{fig:time_}
    \end{subfigure}
    \caption{Validation reward as a function of step (\ref{fig:step_}) and relative time (\ref{fig:time_}). The horizontal dashed line indicates the maximum reward achieved by plain GRPO, and the vertical dashed lines indicate when GRPO with each threshold surpasses the plain GRPO's maximum reward.}
    \label{fig:time}
\end{figure*}

In Figure \ref{fig:difficulty}, we collect the exact batches curated through balanced online difficulty filtering with $(T_\mathrm{Low}, T_\mathrm{High}) = (0.2, 0.8)$ and measure the ``difficulty'' that each model perceives through $1 - p(\mathbf{x})$ for four checkpoints: before, during, and after GRPO, along with the external proxy Qwen2.5-7B-Instruct. As anticipated, the checkpoint evaluated during GRPO maintains an average difficulty of around 0.5, dynamically providing suitably challenging examples throughout the training process. However, both before and after GRPO checkpoints perceive incremental difficulty increases across the curated batches, indicating that the training examples become objectively more challenging over time. Moreover, the external proxy model consistently perceives lower difficulty relative to the initial model but higher difficulty than the final trained model (``After GRPO''). 

This observation, with the results in Table \ref{tab:main_result}, shows that offline difficulty filtering with external proxies can provide partially meaningful difficulty assessments while not being perfectly aligned to the training model's capability, shown through marginal improvements in Table \ref{tab:main_result} compared to plain GRPO. However, the advantage of the balanced online difficulty filtering is still evident in better performance and efficiency.

\subsection{Training efficiency analysis}
Figure \ref{fig:time} illustrates the progression of the reward in the validation set, plotted against both the training steps (\ref{fig:step_}) and the training time on the wall clock (\ref{fig:time_}). As shown in Figure \ref{fig:step_}, models trained with balanced online difficulty filtering consistently outperform the plain GRPO ($0 \leq p(\mathbf{x}) \leq 1$) in fewer training steps. For instance, using $0.4<p(\mathbf{x})<0.6$ achieved the highest overall reward and took the fewest gradient updates to outperform the plain GRPO within less than $50\%$ of steps in Figure \ref{fig:step_}. This suggests that by filtering out less informative examples, the average learnability within each batch increases, allowing faster learning progress. Interestingly, Figure \ref{fig:time_} shows that this benefit carries over even when measured by wall-clock time by exceeding plain GRPO's maximum reward in less training time. However, we also observe that overly aggressive filtering, such as in the case of the $0.4 < p(\mathbf{x}) < 0.6$ setting, can require significantly more rollouts to fill a batch, leading to longer training times overall. These results suggest that online filtering can enable more efficient learning even in real-world settings.

\subsection{Theoretical generalizability}
\label{subsec:general_rewards}

We extend the learnability view of difficulty filtering in reinforcement learning for language models to a broader class of reward distributions. 
We begin by expressing the reverse KL divergence between the initial policy $\pi_{\mathrm{init}}(\cdot)$ and the optimal policy $\pi^*(\cdot)$ as the cumulant generating function (CGF) of a \emph{centered} reward.
We report the detailed proofs for the proposition and corollaries in Appendix \ref{apdx:learnability_general}.

\begin{proposition}[Reverse KL as CGF of centered reward]
\label{prop:cgf_identity}
Fix a prompt $x$ and temperature $\beta>0$. Let $\mu(x):=\mathbb{E}_{y\sim\pi_{\mathrm{init}}(\cdot|x)}[r(x,y)]$. Then
\begin{equation}
\begin{split}
&\mathbb{D}_\mathrm{KL}\!\big(\pi_{\mathrm{init}}(\cdot|x)\,\|\,\pi^*(\cdot|x)\big)
\\&=
\log \,\mathbb{E}_{y\sim\pi_{\mathrm{init}}(\cdot|x)}\left[\exp\!\left(\frac{r(x,y)-\mu(x)}{\beta}\right)\right]
\\&=
K_{\,r(x,\cdot)-\mu(x)}\!\left(\tfrac{1}{\beta}\right),
\label{eq:cgf_identity_main}
\end{split}
\end{equation}
where $K_Z(t):=\log \mathbb{E}[\exp(tZ)]$ denotes the CGF of a random variable $Z$.
\end{proposition}
\noindent \textit{Proof sketch.}
Starting from the soft value $V^*(x)=\beta\log\mathbb{E}[e^{r(x,y)/\beta}]$ and the identity in \eqref{eq:logratio},
\begin{equation}
\begin{split}
&\mathbb{D}_\mathrm{KL}\!\big(\pi_{\mathrm{init}}\|\pi^*\big)
=
-\mathbb{E}\!\left[\log\frac{\pi^*}{\pi_{\mathrm{init}}}\right]
\\&=
\log\mathbb{E}\!\left[e^{r/\beta}\right] - \frac{1}{\beta}\,\mathbb{E}[r]
\\&=
\log\mathbb{E}\!\left[e^{(r-\mu)/\beta}\right],
\end{split}
\end{equation}
and the right-hand side is precisely the CGF of $r-\mu$ at $t=\tfrac{1}{\beta}$. \hfill$\square$

Proposition~\ref{prop:cgf_identity} highlights that learnability quantified by the reverse KL divergence is governed by the \emph{fluctuations} of the reward around its mean. 
The variance term used as a proxy in \S\ref{sec:learnability} is the leading component of this control. 
This CGF view naturally generalizes to two practical settings: (1) \emph{Gaussian rewards}, representing continuous rewards from neural evaluators \citep{lambert2025rewardbench,wen2025rethinking}; and (2) \emph{multinomial rewards}, representing combinations of verifiable objectives \citep{guo2025deepseek,gunjal2025rubricsrewardsreinforcementlearning}.

\begin{corollary}[Gaussian rewards]
\label{cor:gaussian_exact}
If $r_G(x,y)\sim \mathcal{N}(\mu_G(x),\sigma_G^2(x))$ under $\pi_{\mathrm{init}}(\cdot|x)$, then
\begin{equation}
\mathbb{D}_\mathrm{KL}\!\big(\pi_{\mathrm{init}}(\cdot|x)\,\|\,\pi^*(\cdot|x)\big)
\;=\;
\frac{\sigma_G^2(x)}{2\,\beta^2}.
\end{equation}
\end{corollary}

For Gaussian rewards, the reverse KL divergence, \ie the learnability, is \emph{exactly proportional} to the reward variance. 
Consequently, the sample variance from rollouts provides an unbiased and consistent empirical estimate of learnability. Let the sample variance
\begin{equation}
\widehat{\sigma_G^2}(x)\;=\;\frac{1}{n-1}\sum_{i=1}^{n}\big(r_i-\bar r\big)^2,
\end{equation}
where $r_i$ are $n$ independent reward samples. Since $\mathbb{E}[\widehat{\sigma_G^2}(x)]=\sigma_G^2(x)$,
\begin{equation}
\mathbb{E}\!\left[\frac{1}{2\beta^2}\,\widehat{\sigma_G^2}(x)\right]=
\mathbb{D}_\mathrm{KL}\!\big(\pi_{\mathrm{init}}(\cdot|x)\,\|\,\pi^*(\cdot|x)\big).
\end{equation}
Thus, the sample reward variance with sufficient size of rollouts can be a reliable proxy for learnability, even in the Gaussian reward setting, such as the classifier reward models \citep{lambert2025rewardbench}.

\begin{corollary}[Multinomial rewards]
\label{cor:multinomial_lower}
Suppose $r_M(x,y)\in\{0,1,\dots,N\}$ with variance $\sigma^2(x)$ and cumulants $\kappa_k(x)$. Then, for $t=\tfrac{1}{\beta}$,
\begin{equation}
\begin{split}
&\mathbb{D}_\mathrm{KL}\!\big(\pi_{\mathrm{init}}(\cdot|x)\,\|\,\pi^*(\cdot|x)\big)
=
K_{\,r_M-\mu}\!(t)
\\&=
\frac{\sigma^2(x)}{2}\,t^2 + \frac{\kappa_3(x)}{6}\,t^3 +\frac{\kappa_4(x)}{24}\,t^4 +\cdots \\
&\ge
\frac{\sigma^2(x)}{2\,\beta^2}-\mathcal{O}\!\left(\frac{1}{\beta^3}\right),
\end{split}
\end{equation}
and in particular the variance $\tfrac{\sigma^2(x)}{2\beta^2}$ provides a tight second-order lower control of the reverse KL.
\end{corollary}

Here, the leading variance term is the dominant contribution to learnability, while the higher cumulants $\kappa_3,\kappa_4,\dots$ account for skewness and tail behavior typical of multi-objective or rubric-based rewards. The binary case is a special instance with $N{=}1$. If $r_B(x,y)\in\{0,1\}$, then
\begin{equation}
\mathrm{Var}[r_B(x,y)] \;=\; p(x)\big(1-p(x)\big),
\end{equation}
and substituting into the series above recovers the bound used in \S\ref{sec:learnability}:
\begin{equation}
\mathbb{D}_\mathrm{KL}\!\big(\pi_{\mathrm{init}}(\cdot|x)\,\|\,\pi^*(\cdot|x)\big)
\;\ge\;
\frac{p(x)\big(1-p(x)\big)}{2\,\beta^2}.
\end{equation}
This expands the applicability of the learnability viewpoint for the online difficulty filtering to the real-world setting, where multiple verifiable rewards are engaged, \eg format rewards and accuracy rewards in reasoning \citep{guo2025deepseek,deepscaler2025,wei2025redit}.

We further extend the principled discussion on algorithm-agnostic generalizability of the balanced online difficulty filtering in Appendix \ref{apdx:discussion}.

%% file: sections/8_conclusion.tex
\section{Conclusion}

In this work, we established the theoretical foundations of online difficulty-aware filtering in reinforcement learning with verifiable rewards (RLVR). Our analysis showed that tasks with intermediate difficulty maximize the lower bound of policy improvement, providing a principled explanation for the effectiveness of difficulty-based data curation, which was heuristically adopted in previous works. Through extensive ablation studies across multiple reasoning benchmarks, we verified these theoretical insights and demonstrated consistent gains in both sample efficiency and performance. These results highlight the importance of theoretically grounded difficulty control in RLVR, offering a unified perspective that connects empirical heuristics with formal learning principles.





%% file: sections/limitations.tex
\section*{Limitations}

Our work provides both theoretical and empirical guidelines for online difficulty filtering in reasoning-oriented reinforcement learning for language models. While our theoretical analysis can be applied to any verifiable task, as we have shown through the proposition and corollaries, our empirical validation was conducted solely on math reasoning tasks. We leave the exploration of diverse verifiable tasks, such as coding and scientific reasoning, for future work. Furthermore, we plan to investigate the broader applicability of our method to larger scales and wider model families.

%% file: sections/apdx.tex
\onecolumn

\appendix

\section{Related Works}\label{apdx:related}

\paragraph{Reinforcement learning with verifiable rewards.} Recent advancements demonstrate significant reasoning improvements in LLMs through RL \citep{havrilla2024teaching, openai2024openaio1card, lambert2024t, guo2025deepseek, olmo20252olmo2furious, kumar2025training}. 
OpenAI o1 \citep{openai2024openaio1card} initially reported that increasing the compute during RL training and inference improves reasoning performance. DeepSeek R1 \citep{guo2025deepseek} further found that, in reinforcement learning with verifiable rewards (RLVR), longer responses correlate with better reasoning.
Concurrent studies \citep{team2025kimi,hou2025advancing,deepscaler2025} employed algorithms, such as GRPO \citep{shao2024deepseekmath} or RLOO \citep{ahmadian2024back}, relying on advantage estimation via sampling rather than PPO-like value networks. \citet{hou2025advancing} further found that training efficiency improved with increased sampling in RLOO, invoking the need for more sample-efficient training strategies in RLVR.

\paragraph{Difficulty-based curriculum learning.} Curriculum learning has been widely adopted in fine-tuning LLMs to improve training efficiency \citep{lee2024instruction, nair2024curriculum, team2025kimi, cui2025process}. 
Static curricula, \ie offline data curation with a predetermined task difficulty, have been effective in multiple domains: instruction-tuning \citep{lee2024instruction} and coding \citep{nair2024curriculum,team2025kimi, li2025limr} to name a few. In RLVR, \citet{team2025kimi} employs a static difficulty-based curriculum, assigning tasks at fixed difficulty levels to ensure efficient progression. Similarly, \citet{li2025limr} selects a high-impact subset of training data based on a ``learning impact measure''. 
Meantime, adaptive curricula dynamically adjust task difficulty based on the learners' progress, addressing the limitations of static curricula \citep{florensa2018automatic,cui2025process}. 
Specifically, \citet{cui2025process} applied adaptive filtering in reasoning and reported an empirical advantage in reducing reward variance. However, \citet{meng2025mm} observed that such dynamic exclusion of examples may destabilize training, as it causes fluctuations in the effective batch size.

These works introduce adaptive filtering heuristics for RLVR but do not characterize \textit{when} and \textit{why} such filtering helps or how it interacts with KL-regularized RL objectives. In contrast, our work (1) derives a reverse-KL lower bound that identifies reward variance as the key learnability proxy, (2) shows that this directly motivates intermediate-difficulty selection, and (3) proposes an asynchronous fixed-batch implementation that avoids the batch-size instability.

\clearpage

\section{Asynchronous Implementation of Online Difficulty Filtering}\label{apdx:implementation}

We provide a detailed diagram depicting the practical implementation of the online difficulty filtering, especially with the asynchronous setting \citep{noukhovitch2025faster}. The formal expression of filling the batch $\mathcal{B}^{(t)}$ for the balanced online difficulty filtering is:
\begin{equation}
    \mathcal{B}^{(t)} = \left\{ \left(\mathbf{x}, \{\mathbf{y}_i, r_\mathrm{acc}(\mathbf{x}, \mathbf{y}_i)\}_{i=1}^{G}\right) \mid T_{\text{Low}} \le \frac{1}{G} \sum_{i=1}^{G} r_\mathrm{acc}(\mathbf{x}, \mathbf{y}_i) \le T_{\text{High}},~\mathbf{y}_i \sim \pi_{\theta_t}(\cdot|\mathbf{x}) \right\}.\label{eq:batch}
\end{equation}
Here, the sample mean of $r_{acc}(\mathbf{x}, \mathbf{y}_i)$ is an unbiased estimate of $\mathbb{E}_{\mathbf{y}\sim \pi_{\theta_t}(\cdot|\mathbf{x})}\left[ r_\mathrm{acc}(\mathbf{x}, \mathbf{y}) \right]$. Rollouts for each prompt are sampled asynchronously and in parallel, enabling continuous batching of prompts and rollouts \citep{daniel2023continuous, kwon2023efficient, noukhovitch2025faster}. Each prompt’s visit count, $\mathrm{vc}(\mathbf{x})$, is incremented after generating $G$ rollouts, ensuring it isn’t re-processed in the same iteration. Moreover, the active rollout process $\mathcal{P}_{\text{active}}$ is halted once the batch capacity is reached, allowing prompt training with the collected data. This sampling-based framework is compatible with Monte Carlo methods such as RLOO \citep{ahmadian2024back} and VinePPO \citep{kazemnejad2024vineppo}. 
\begin{figure}[hb!]
    \centering
    \vspace{-0.05in}
    \includegraphics[width=0.8\columnwidth]{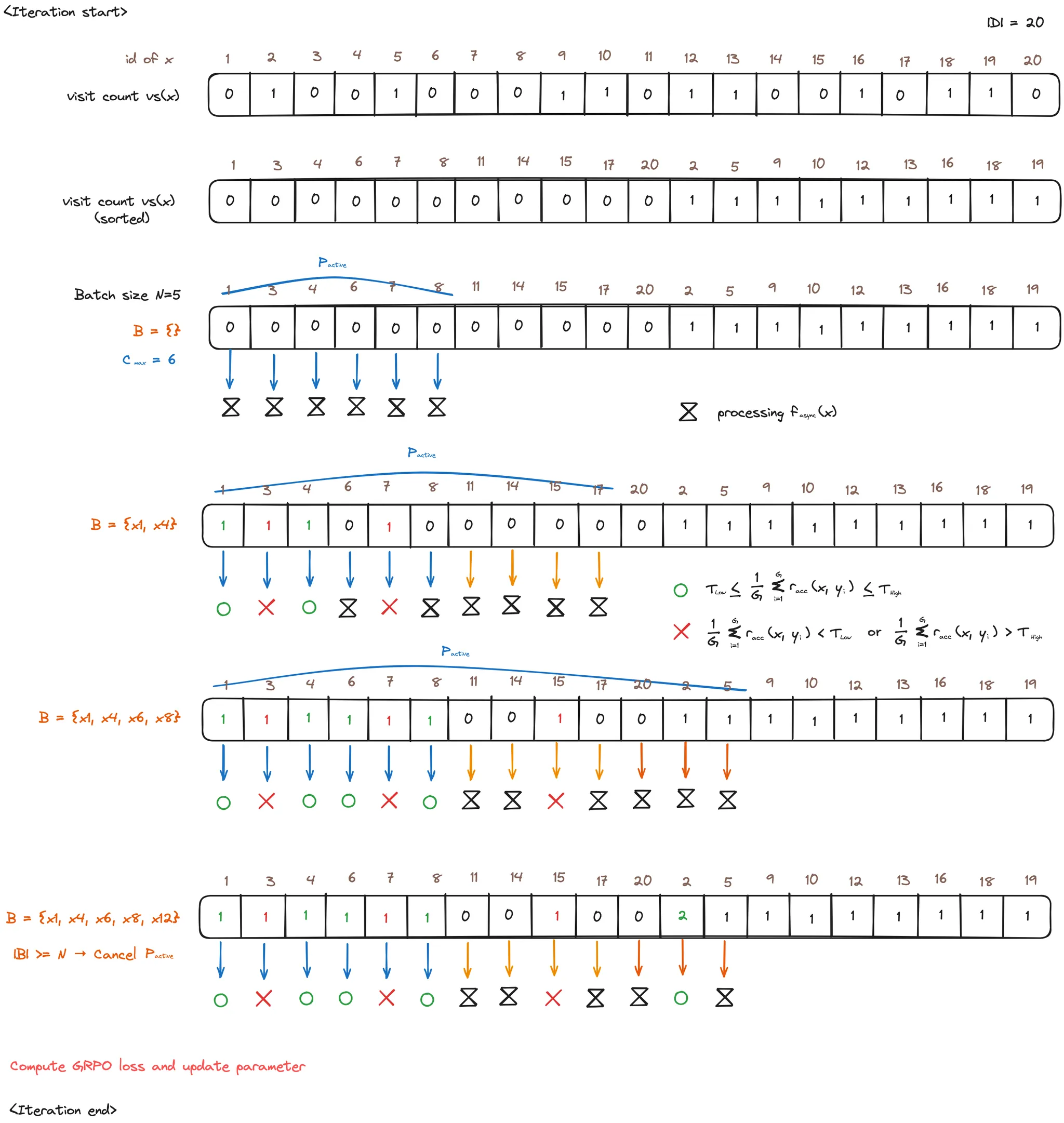}
    \caption{Illustration of the rollout process in the proposed algorithm with online difficulty filtering. Each iteration begins by sorting the dataset based on the visit count $\mathrm{vc(\mathbf{x})}$ of each example $\mathbf{x}$. A batch of unvisited or least-visited prompts is selected, respecting a predefined concurrency limit $C_{\text{max}}$. The asynchronous function $f_{async}$ samples responses from the current policy and evaluates them using the accuracy reward $r_{acc}$. Prompts with a pass rate within the accepted range $[T_{\text{Low}}, T_{\text{High}}]$ are added to the training batch. Once the batch $\mathcal{B}$ reaches the target size $N$, any remaining asynchronous jobs in $\mathcal{P}_{\text{active}}$ are canceled. The policy is then updated using the GRPO loss computed over the collected batch.}
    \label{fig:high_level}
\end{figure}

\clearpage

\section{Learnability in Soft Prompts}
\label{apdx:learnability}

We prove the variance-controlled lower bound in Proposition~\ref{prop:variance_controlled}. 

\textit{Proof.} Throughout, let $r_{\mathrm{acc}}(x,y)\in\{0,1\}$ denote a binary verifiable reward with
\begin{equation}
P\!\big(r_{\mathrm{acc}}(x,y)=1\big)=p(x)~~\text{and}~~P\!\big(r_{\mathrm{acc}}(x,y)=0\big)=1-p(x),
\end{equation}
and assume $\beta>0$ with $1/\beta\ll 1$ so that second-order Taylor expansions are valid. Expectations are taken with respect to $y\sim\pi_{\mathrm{init}}(\cdot\mid x)$ unless noted.

\paragraph{Step 1: Soft value under Bernoulli rewards.}
Define the random variable
\begin{equation}
Y \;=\; \exp\!\Big(\tfrac{1}{\beta}\,r_{\mathrm{acc}}(x,y)\Big)
\;=\;
\begin{cases}
1, & r_{\mathrm{acc}}(x,y)=0,\\
\exp(1/\beta), & r_{\mathrm{acc}}(x,y)=1.
\end{cases}
\end{equation}
Then
\begin{equation}
\mathbb{E}[Y] \;=\; (1-p(x)) + p(x)\,e^{1/\beta}.
\end{equation}
Using the soft value definition, we obtain
\begin{equation}
V^*(x) \;=\; \beta \,\log\Big((1-p(x)) + p(x)\,e^{1/\beta}\Big).
\label{eq:apdx_Vstar}
\end{equation}
~~

\paragraph{Step 2: Expected log-ratio.}
From the identity in \eqref{eq:logratio}, substituting $\mathbb{E}[r_{\mathrm{acc}}(x,y)]=p(x)$ and \eqref{eq:apdx_Vstar} gives
\begin{equation}
\mathbb{E}\!\left[\log\frac{\pi^*(y\mid x)}{\pi_{\mathrm{init}}(y\mid x)}\right]
\;=\;
\frac{p(x)}{\beta}
-
\log\!\Big((1-p(x)) + p(x)\,e^{1/\beta}\Big).
\label{eq:apdx_logratio_diff}
\end{equation}
~~

\paragraph{Step 3: Second-order expansion and lower bound.}
Write $e^{1/\beta}=1+\tfrac{1}{\beta}+\tfrac{1}{2\beta^2}+\mathcal{O}(\beta^{-3})$. Then
\begin{equation}
(1-p(x)) + p(x)\,e^{1/\beta}
\;=\;
1 + p(x)\!\left(\frac{1}{\beta}+\frac{1}{2\beta^2}\right) + \mathcal{O}\!\left(\frac{1}{\beta^3}\right).
\end{equation}
Using $\log(1+\epsilon)\ge \epsilon - \tfrac{\epsilon^2}{2}$ with $\epsilon=p(x)\!\left(\tfrac{1}{\beta}+\tfrac{1}{2\beta^2}\right)$ yields
\begin{equation}
\log\!\Big((1-p(x)) + p(x)\,e^{1/\beta}\Big)
\;\ge\;
\frac{p(x)}{\beta}
+
\frac{p(x)\big(1-p(x)\big)}{2\beta^2}
+\mathcal{O}\!\left(\frac{1}{\beta^3}\right).
\label{eq:apdx_log_lower}
\end{equation}
Subtracting \eqref{eq:apdx_log_lower} from \eqref{eq:apdx_logratio_diff} gives
\begin{equation}
\mathbb{E}\!\left[\log\frac{\pi^*(y\mid x)}{\pi_{\mathrm{init}}(y\mid x)}\right]
\;\le\;
-\frac{p(x)\big(1-p(x)\big)}{2\beta^2}
+\mathcal{O}\!\left(\frac{1}{\beta^3}\right).
\end{equation}
By the reverse KL identity in \eqref{eq:logratio},
\begin{equation}
D_{\mathrm{KL}}\!\big(\pi_{\mathrm{init}}(\cdot\mid x)\,\|\,\pi^*(\cdot\mid x)\big)
\;=\;
-\mathbb{E}\!\left[\log\frac{\pi^*(y\mid x)}{\pi_{\mathrm{init}}(y\mid x)}\right]
\;\ge\;
\frac{p(x)\big(1-p(x)\big)}{2\beta^2}
+\mathcal{O}\!\left(\frac{1}{\beta^3}\right).
\end{equation}
Dropping higher-order terms establishes the stated bound:
\begin{equation}
D_{\mathrm{KL}}\!\big(\pi_{\mathrm{init}}(\cdot\mid x)\,\|\,\pi^*(\cdot\mid x)\big)
\;\ge\;
\frac{p(x)\big(1-p(x)\big)}{2\beta^2}.
\end{equation}
~~

\paragraph{Extremal cases.}
If $p(x)\in\{0,1\}$, then $r_{\mathrm{acc}}(x,y)$ is almost surely constant under $\pi_{\mathrm{init}}$, so $V^*(x)=\mathbb{E}[r_{\mathrm{acc}}(x,y)]$ and \eqref{eq:apdx_logratio_diff} evaluates to $0$. Hence
\begin{equation}
D_{\mathrm{KL}}\!\big(\pi_{\mathrm{init}}(\cdot\mid x)\,\|\,\pi^*(\cdot\mid x)\big)=0,
\end{equation}
which matches Proposition~\ref{prop:variance_controlled} at the endpoints.

\hfill$\square$

\section{Training Configurations}\label{apdx:config}

All experiments are built on the Qwen2.5-3B base model \citep{yang2024qwen2}. We integrate DeepSpeed ZeRO-3 \citep{rajbhandari2020zero} optimization in our training pipeline to handle memory and computation efficiently. Both the SFT and RLVR stages are conducted on a distributed setup of 8$\times$NVIDIA A100 (80GB) GPUs.

\paragraph{Training Data Curation} For SFT, we sample problems from the NuminaMath dataset \citep{li2024numinamath} and generate solutions using DeepSeek-R1 \citep{guo2025deepseek}. Only samples with verifiably correct solutions are retained, and we stop once approximately 1,000 such problem-solution pairs are collected. The final SFT dataset contains 1,107 filtered problems. For RLVR, we adopt a subset of the public dataset used in \citet{cui2025process}\footnote{https://huggingface.co/datasets/PRIME-RL/Eurus-2-RL-Data}. We specifically use only the math domain problems. This dataset provides a diverse pool of challenging prompts.
\paragraph{Supervised fine-tuning} We use a learning rate of $5 \times 10^{-6}$ and fine-tune it for 5 epochs. The learning rate schedule is linear, with the first 25 steps used for warm-up. We use a batch size of 21.
\paragraph{Reinforcement learning} We utilize the SGLang \citep{zheng2025sglang} framework to accelerate parallel rollout generation, enabling efficient sampling of multiple reasoning trajectories. Training is run for 256 steps, with empirical performance gains saturating after roughly 128 steps. Each update uses 16 sampled rollouts with 16 distinct prompts per batch, followed by a one-step policy update per rollout.

\paragraph{Reward design} To guide the model toward producing responses aligned with the DeepSeek R1 format, we introduce a \textbf{format reward} based on five constraints: (1) the response must begin with a `$<$think$>$' tag, (2) the `$<$think$>$' section must be properly closed with a `$<$/think$>$' tag, (3) the `$<$think$>$' section must be non-empty, (4) the summary section following `$<$/think$>$' must also be non-empty, and (5) the response must terminate with an eot token. Each constraint contributes 0.2 points, resulting in a maximum format reward of 1.0. In addition, we implement a \textbf{language reward} to reduce language mixing, especially given that all prompts during training and evaluation are in English. This reward was computed as the ratio of characters in the response that are alphabetic, symbolic (e.g., mathematical symbols), or whitespace, and ranged from 0 to 1. Lastly, we define an \textbf{accuracy reward}, assigning a score of 1.0 for correct answers and 0.0 for incorrect ones. The total reward is the sum of these three components—format, language, and accuracy—yielding a final reward score between 0 and 3.

\section{Evaluation Benchmarks}\label{apdx:eval_benchmarks}
We employ five different challenging math reasoning benchmarks:
\begin{itemize}[left=2pt]
  \item \textbf{MATH500} \citep{hendrycks2measuring} consists of 500 problems sampled from \citet{lightman2023let}, maintaining topic and difficulty balance.
  \item \textbf{AIME} \citep[American Invitational Mathematics Examination]{li2024numinamath} uses 30 problems from the 2024 official competition, while \textbf{AMC} \citep[American Mathematics Competitions]{li2024numinamath} includes 40 problems from the 2023 official competition. Both benchmarks consist of contest-level advanced mathematical problems.
  \item \textbf{MinervaMath} \citep{lewkowycz2022solving} evaluates quantitative reasoning with complex mathematical problems at an undergraduate or Olympiad level.
  \item \textbf{OlympiadBench} \citep{he2024olympiadbench} includes 674 open-ended text-only competition problems from a broader set of 8,476 Olympiad and entrance exam questions, specifically using the ``OE\_TO\_maths\_en\_COMP'' subset.
\end{itemize}
Inference is conducted via SGLang \citep{zheng2025sglang} with top-$p$ set to 0.95, temperature set to 0.6, and the maximum number of output tokens limited to 8,192.

\section{Difficulty-Aware Performance Analysis}\label{apdx:difficulty}
To further understand the effect of our method, we analyze performance variations based on difficulty levels.

\paragraph{Benchmark-Level Difficulty Spectrum}
As discussed in \S~\ref{sec:exp}, our benchmark suite spans a wide difficulty range. This is reflected in the SFT checkpoint performance of Qwen2.5-3B, which ranges from 0.0\% to 49.8\% as shown in Table~\ref{tab:main_result}. We order the benchmarks in ascending difficulty according to SFT performance: AIME (0.0\%), MinervaMath (13.2\%), OlympiadBench (17.3\%), AMC (20.5\%), and MATH500 (49.8\%).
From this perspective, we observe two trends:
\begin{itemize}
    \item Narrowing the difficulty threshold (i.e., tighter filtering range) generally improves performance, especially on challenging tasks like MinervaMath and AIME.
    \item Harder benchmarks benefit more from filtering. For instance, AIME shows more than a 300\% relative improvement over SFT, and MinervaMath improves by 35\%.
\end{itemize}

\paragraph{Difficulty-Level Breakdown within MATH500}
We also analyze performance by difficulty levels in the MATH500 benchmark. Table~\ref{tab:math500_difficulty} shows that balanced filtering GRPO outperforms plain GRPO across most difficulty levels, especially on harder ones (Level 3–5).
\input{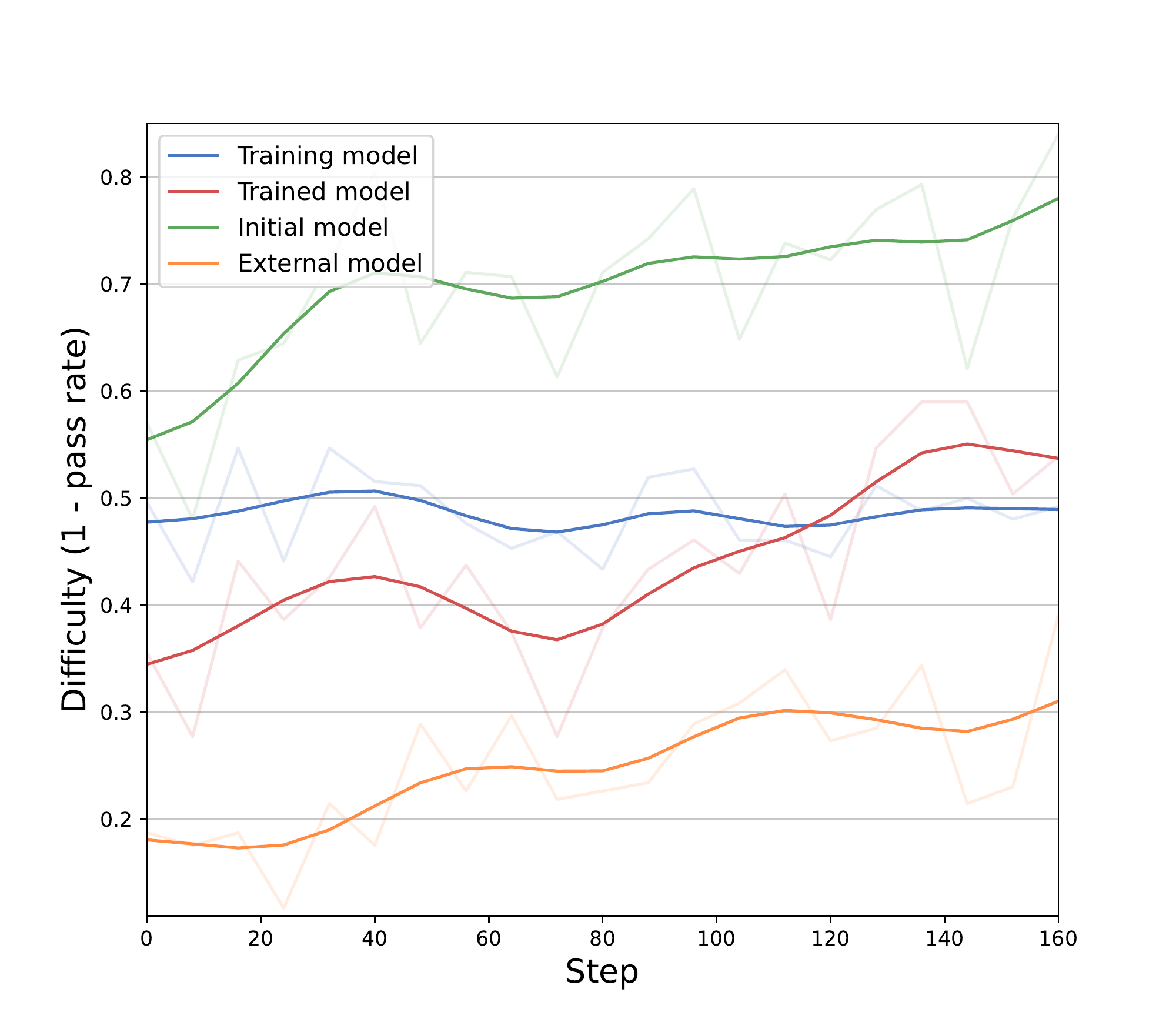}

\section{General rewards: CGF derivations and bounds}
\label{apdx:learnability_general}

We provide full proofs for Proposition~\ref{prop:cgf_identity}, Corollary~\ref{cor:gaussian_exact}, and Corollary~\ref{cor:multinomial_lower}.

\subsection{Proof of Proposition~\ref{prop:cgf_identity}.}\label{apdx:prop_cgf}

\textit{Proof.} Fix $x$ and write expectations over $y\sim\pi_{\mathrm{init}}(\cdot|x)$. By the definition of the soft value,
\begin{equation}
V^*(x) \;=\; \beta \,\log \mathbb{E}\!\left[\exp\!\left(\tfrac{1}{\beta}\,r(x,y)\right)\right].
\end{equation}
From \eqref{eq:logratio},
\begin{equation}
\mathbb{E}\!\left[\log\frac{\pi^*(y|x)}{\pi_{\mathrm{init}}(y|x)}\right]
\;=\;
\frac{1}{\beta}\,\mathbb{E}[r(x,y)] \;-\; \frac{1}{\beta}\,V^*(x)
\;=\;
\frac{1}{\beta}\,\mu(x) \;-\; \log \mathbb{E}\!\left[e^{r/\beta}\right].
\end{equation}
Multiplying both sides by $-1$ gives
\begin{equation}
D_{\mathrm{KL}}\!\big(\pi_{\mathrm{init}}\|\pi^*\big)
\;=\;
-\,\mathbb{E}\!\left[\log\frac{\pi^*}{\pi_{\mathrm{init}}}\right]
\;=\;
\log \mathbb{E}\!\left[e^{r/\beta}\right] \;-\; \frac{1}{\beta}\,\mu(x).
\end{equation}
Factor out the mean inside the exponential:
\begin{equation}
\log \mathbb{E}\!\left[e^{r/\beta}\right]
\;=\;
\log \mathbb{E}\!\left[e^{(r-\mu)/\beta}\,e^{\mu/\beta}\right]
\;=\;
\frac{\mu}{\beta} \;+\; \log \mathbb{E}\!\left[e^{(r-\mu)/\beta}\right].
\end{equation}
Substitute into the previous line to obtain
\begin{equation}
D_{\mathrm{KL}}\!\big(\pi_{\mathrm{init}}\|\pi^*\big)
\;=\;
\log \mathbb{E}\!\left[\exp\!\left(\tfrac{r-\mu}{\beta}\right)\right]
\;=\;
K_{\,r-\mu}\!\left(\tfrac{1}{\beta}\right),
\end{equation}
which is the desired identity. \hfill$\square$

\subsection{Proof of Corollary~\ref{cor:gaussian_exact}.}\label{apdx:for_gaussian}
\textit{Proof.} Let $Z:=r_G-\mu_G$. Under $\pi_{\mathrm{init}}$, $Z\sim \mathcal{N}(0,\sigma_G^2)$, whose moment generating function is $\mathbb{E}[e^{tZ}]=\exp(\tfrac{1}{2}\sigma_G^2 t^2)$. Therefore the CGF is $K_Z(t)=\tfrac{1}{2}\sigma_G^2 t^2$. Applying Proposition~\ref{prop:cgf_identity} with $t=\tfrac{1}{\beta}$ yields
\begin{equation}
D_{\mathrm{KL}}\!\big(\pi_{\mathrm{init}}\|\pi^*\big)
\;=\;
K_Z\!\left(\tfrac{1}{\beta}\right)
\;=\;
\frac{\sigma_G^2}{2\,\beta^2}.
\end{equation}
\hfill$\square$

\subsection{Proof of Corollary~\ref{cor:multinomial_lower}.}\label{apdx:for_multi}
\textit{Proof.} Let $Z:=r_M-\mu$. Write the CGF Taylor series around $t=0$:
\begin{equation}
K_Z(t) \;=\; \sum_{k=2}^{\infty} \frac{\kappa_k}{k!}\,t^k
\;=\;
\frac{\kappa_2}{2}\,t^2 \;+\; \frac{\kappa_3}{6}\,t^3 \;+\; \frac{\kappa_4}{24}\,t^4 \;+\; \cdots,
\end{equation}
where $\kappa_2=\sigma^2=\mathrm{Var}(Z)$ and $\kappa_k$ are the cumulants of $Z$. Because $r_M$ is bounded (multinomial or $[a,b]$), all moments of $Z$ are finite and thus all $\kappa_k$ are finite. For $t=\tfrac{1}{\beta}$,
\begin{equation}
K_Z\!\left(\tfrac{1}{\beta}\right)
\;=\;
\frac{\sigma^2}{2\,\beta^2}
\;+\;
\frac{\kappa_3}{6\,\beta^3}
\;+\;
\frac{\kappa_4}{24\,\beta^4}
\;+\;\cdots.
\end{equation}
Hence,
\begin{equation}
K_Z\!\left(\tfrac{1}{\beta}\right)
\;\ge\;
\frac{\sigma^2}{2\,\beta^2}
\;-\;
\left|\frac{\kappa_3}{6\,\beta^3}\right|
\;-\;
\left|\frac{\kappa_4}{24\,\beta^4}\right|
\;-\;\cdots,
\end{equation}
which shows that the variance term dominates at second order and provides a lower control up to $\mathcal{O}(\beta^{-3})$. In particular, for sufficiently large $\beta$ (or small $t$), we obtain
\begin{equation}
K_Z\!\left(\tfrac{1}{\beta}\right)
\;\ge\;
\frac{\sigma^2}{2\,\beta^2}
\;-\;
\mathcal{O}\!\left(\frac{1}{\beta^3}\right).
\end{equation}
Finally, apply Proposition~\ref{prop:cgf_identity} to translate this bound to the reverse KL, completing the proof. \hfill$\square$

\clearpage

\section{Discussion: Native Generalizability of Online Difficulty Filtering in RL for LLMs}
\label{subsec:discussion_generalizability}\label{apdx:discussion}

The concept of online difficulty filtering in reinforcement learning for language models is \textbf{inherently versatile across verifiable tasks and reward formulations}, even beyond the empirical scope of this paper. 
Recent work in reinforcement learning with verifiable rewards (RLVR) demonstrates its applicability to diverse domains, including logic puzzles \citep{chen2025enigmatascalinglogicalreasoning}, medical reasoning \citep{gunjal2025rubricsrewardsreinforcementlearning}, complex instruction-following \citep{pyatkin2025generalizingverifiableinstructionfollowing}, and coding \citep{chen2025r1codeinterpreterllmsreasoncode}.
Our theoretical framework builds on the standard KL-regularized or soft reinforcement learning formulation \citep{schulman2018equivalencepolicygradientssoft,richemond2024offlineregularisedreinforcementlearning}, which optimizes the objective
\begin{equation}
\max_{\pi_\theta}\; \mathbb{E}_{y\sim \pi_\theta(\cdot|x)}[\,r(x,y)\,]
 - \beta\,\mathrm{D}_{\mathrm{KL}}\!\big(\pi_\theta(\cdot|x)\,\|\,\pi_{\mathrm{init}}(\cdot|x)\big),
\end{equation}
yielding the Boltzmann-rational optimal policy relative to the reference $\pi_{\mathrm{init}}$. Modern post-training algorithms such as PPO \citep{schulman2017proximal}, GRPO \citep{shao2024deepseekmath}, RLOO, and REINFORCE$++$ \citep{hu2025reinforceefficientrlhfalgorithm} can all be interpreted as optimizing surrogates of this KL-regularized objective, differing primarily in how the policy gradient is estimated or regularized. Hence, the theoretical mechanism behind online difficulty filtering, namely that the reverse KL divergence between $\pi_{\mathrm{init}}$ and $\pi^*$ is determined by the cumulant generating function of the centered reward and dominated by its variance, extends directly to these algorithms as long as a KL penalty to the reference is preserved.

The generalized analysis in \S\ref{subsec:general_rewards} further shows that this relationship holds exactly for Gaussian rewards and forms a tight second-order lower bound for multinomial or bounded rewards.  Since continuous neural reward models are approximately Gaussian and rubric-based verifiable rewards are discrete and bounded, these two cases jointly cover most practical reward systems in RLHF and RLVR. Accordingly, the same variance-guided filtering principle applies regardless of task or reward type: prompts with intermediate success probabilities maximize the effective learning signal, while trivially easy or hard ones contribute negligible gradients. This makes balanced online difficulty filtering a theoretically grounded and natively general component for PPO-, GRPO-, or RLOO-style alignment training across diverse verifiable tasks.





%% file: table/difficulty.tex
\begin{table}[h]
\centering
\footnotesize
\begin{tabular}{lccc}
\toprule
\textbf{Difficulty} & \textbf{Plain ($0 \leq p(\mathbf{x}) \leq 1$ )} & \textbf{w/ Online Filtering ($0 < p(\mathbf{x}) < 1$)} & \textbf{w/ Online Filtering ($0.3 < p(\mathbf{x}) < 0.7$)} \\
\midrule
Level 1 & \textbf{88.37} & \textbf{88.37} & 83.72 \\
Level 2 & 78.89 & \textbf{83.33} & \textbf{83.33} \\
Level 3 & 71.43 & 70.48 & \textbf{79.05} \\
Level 4 & 47.66 & 50.78 & \textbf{55.47} \\
Level 5 & 30.60 & \textbf{32.84} & 32.09 \\
\bottomrule
\end{tabular}
\caption{Accuracy (\%) of GRPO-trained models on MATH500 by difficulty level. The highest score for each level is in \textbf{bold}.}
\label{tab:math500_difficulty}
\end{table}